\documentclass{sig-alternate}
\usepackage{times}
\usepackage{graphicx}
\usepackage{amsmath}
\usepackage{amssymb}
\usepackage{array}
\usepackage{booktabs}
\usepackage{mathrsfs}
\usepackage{algorithm}
\usepackage{algorithmic}
\usepackage{multirow}
\usepackage{mathtools}

\newcommand{\INPUT}{\item[\myinput]}
\newcommand{\OUTPUT}{\item[\myoutput]}
\newcommand{\myinput}{\textbf{Input:}}
\newcommand{\myoutput}{\textbf{Output:}}
\newcommand{\INITIAL}{\item[\myinitial]}
\newcommand{\myinitial}{\textbf{Initialization:}}
\newcommand{\INTERATION}{\item[\myiter]}
\newcommand{\myiter}{\textbf{Iteration:}}

\newcommand{\MYWHILE}{\item[\mywhile]}
\newcommand{\mywhile}{\textbf{repeat}}

\newcommand{\MYENDWHILE}{\item[\myendwhile]}
\newcommand{\myendwhile}{\textbf{until}}
\begin{document}
%
% --- Author Metadata here ---
%\conferenceinfo{WOODSTOCK}{'97 El Paso, Texas USA}
\CopyrightYear{2013} % Allows default copyright year (20XX) to be over-ridden - IF NEED BE.
\crdata{0-89791-88-6/97/05}  % Allows default copyright data (0-89791-88-6/97/05) to be over-ridden - IF NEED BE.
% --- End of Author Metadata ---

\title{Learning Latent Spatio-Temporal Compositional Model for Human Action Recognition}
%\subtitle{[Extended Abstract]
%\titlenote{A full version of this paper is available as
%\textit{Author's Guide to Preparing ACM SIG Proceedings Using
%\LaTeX$2_\epsilon$\ and BibTeX} at

\numberofauthors{3} %  in this sample file, there are a *total*
\author{
% 1st. author%
\alignauthor
Xiaodan Liang\\
\affaddr{Sun Yat-Sen University, China}\\
\email{xdliang328@gmail.com}\\
\alignauthor
Liang Lin\thanks{Corresponding author is Liang Lin. This work
was supported by the 863 Program of China (no.2012AA011504), the
National Natural Science Foundation of China (no. 61173082), and the
Special Project on the Integration of Industry, Education and Research of
Guangdong Province (no. 2012B091100148), and the Guangdong Science
and Technology Program (no. 2012B031500006). }\\
\affaddr{Sun Yat-Sen University, China}\\
\email{linliang@ieee.org}\\
\alignauthor
Liangliang Cao\\
\affaddr{IBM Research, U.S.}\\
\email{liangliang.cao@us.ibm.com}\\
}
%% 2nd. author
%\alignauthor
%G.K.M. Tobin\titlenote{The secretary disavows
%any knowledge of this author's actions.}\\
%       \affaddr{Institute for Clarity in Documentation}\\
%       \affaddr{P.O. Box 1212}\\
%       \affaddr{Dublin, Ohio 43017-6221}\\
%       \email{webmaster@marysville-ohio.com}

% There's nothing stopping you putting the seventh, eighth, etc.
% author on the opening page (as the 'third row') but we ask,
% for aesthetic reasons that you place these 'additional authors'
% in the \additional authors block, viz.
% Just remember to make sure that the TOTAL number of authors
% is the number that will appear on the first page PLUS the
% number that will appear in the \additionalauthors section.

\maketitle

\begin{abstract}
\vspace{-1mm}
Action recognition is an important problem in multimedia understanding. This paper addresses this problem by building an expressive compositional action model. We model one action instance in the video with an ensemble of spatio-temporal compositions: a number of discrete temporal anchor frames, each of which is further decomposed to a layout of deformable parts. In this way, our model can identify a Spatio-Temporal And-Or Graph (STAOG)  to represent the latent structure of actions \emph{e.g.} triple jumping, swinging and high jumping. The STAOG model comprises four layers: (i) a batch of leaf-nodes in bottom for detecting various action parts within video patches; (ii) the or-nodes over bottom, \emph{i.e.} switch variables to activate their children leaf-nodes for structural variability; (iii) the and-nodes within an anchor frame for verifying spatial composition; and (iv) the root-node at top for aggregating scores over temporal anchor frames. Moreover, the contextual interactions are defined between leaf-nodes in both spatial and temporal domains. For model training, we develop a novel weakly supervised learning algorithm which iteratively determines the structural configuration (\emph{e.g.} the production of leaf-nodes associated with the or-nodes) along with the optimization of multi-layer parameters. By fully exploiting spatio-temporal compositions and interactions, our approach handles well large intra-class action variance (\emph{e.g.} different views, individual appearances, spatio-temporal structures). The experimental results on the challenging databases demonstrate superior performance of our approach over other competing methods.
\end{abstract}
\vspace{-4mm}
\vspace{-2mm}
\vspace{-2mm}

\section{Introduction}

With the popularity of personal video cameras and multi-view video capturing devices, we are entering an era with rich amount of multimedia documents surrounding us. To interact with these videos, there have been increasing demands of understanding human activities in these videos. Although many research studies \cite{SIFTBag,randomforest,fusion,zhu09,MultimediaEvent12,CrossDataset,SIFT,Featurecorrelation12,MultiView,journals/tmm/DuanLC13} have been carried out to understand and retrieve large scale video contents, these works focus on high level semantics instead of describing human activities. There still exists a need to recognize fine-grained information for human activities, \emph{e.g.} body poses and temporal motions.

This paper targets on the challenge of understanding spatial and temporal variances of video phenomenons. More specifically, we are considering the following difficulties:

%With the development of multimedia processing technology, recognizing human actions in videos has received increasing attention in recent years~\cite{MultiView,CrossDataset,SpatioTempGraph,Featurecorrelation12,SIFTBag,randomforest,fusion}. One of the key challenges lies in spatio-temporal variance of video phenomenon. More specifically, the task of human action recognition involves the following difficulties:
\vspace{-2mm}
\begin{itemize}
\setlength{\itemsep}{1pt}
\setlength{\parskip}{0pt}
 \setlength{\parsep}{1pt}
  \item A human body is composed of multiple parts, and different parts are associated with different motions.
  \item Both the short term and long term motions may be fused by different background movement or camera motion, which brings the difficulties of accurately modeling the temporal characteristics for actions.
  \item In real world videos, human actions often happen with uncertainties: some happen along with occlusions, some are caused by view/pose variance, or due to diverse actor appearances and motions.
\end{itemize}
%actioncontext,Laptevbmvc
Due to these difficulties, it is crucial to reduce the spatio-temporal ambiguity when modeling the human actions. Most of the previous studies were built on simplified action models while overlooking the detailed spatio-temporal structure information. Of these works, a large amount of studies were based on spatio-temporal interest points \cite{STIP,YuanLW11,SongTZCZL12}. Some researchers proposed to enrich the action model with the appearance information or context information \cite{ActionTemplate09,Shuicheng09}. Some other researchers learned temporal structures for action recognition \cite{SatkinH10,LatentTem_CVPR2012}. However, none of these works provides an effective model which can unify spatial and temporal information to infer the structure of human motion.
%~\cite{LiuMultiple,LiusemanticFeature}

We aims to develop an effective configurable model, namely the Spatio-Temporal And-Or Graph (STAOG) for action recognition, which addresses the problems mentioned above. Our idea is partially motivated by the image grammar model~\cite{ZhuM06}, which hierarchically decomposes an image pattern with mixed and-nodes and or-nodes, as well as modeling rich structural variations of parts.

The challenges of generalizing And-Or graphs for action recognition are two-folds. First, the traditional And-Or graphs are limited in modeling the hierarchical configuration of spatio-temporal information. Because actions in videos are often more complicated than images, we need more powerful models for video problems. Second, videos require more efficient models that can be effectively learned from large amount of video information without elaborate supervision and initialization.

\begin{figure}[ptb]
\begin{center}
\includegraphics[width=3.5in]{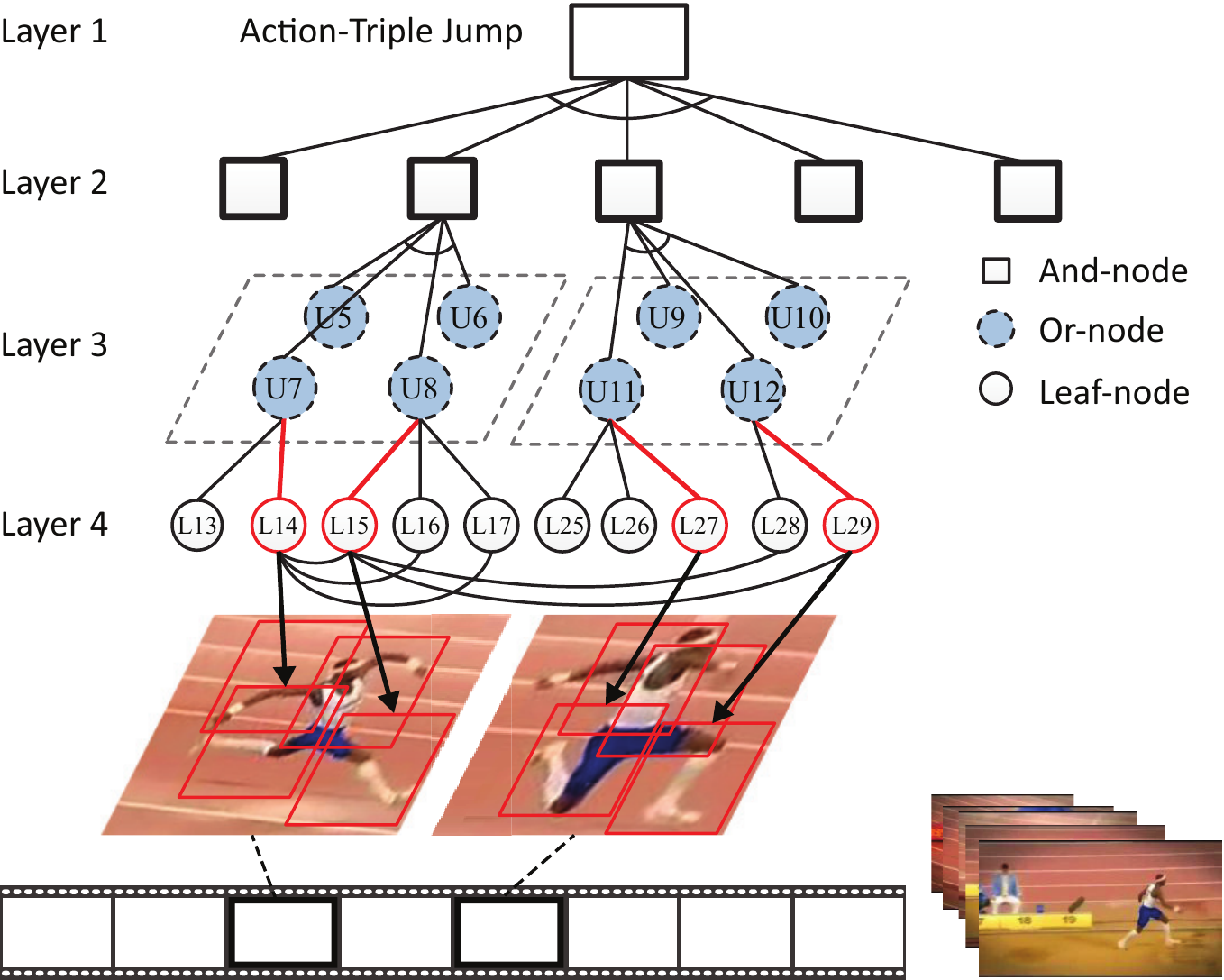}
\end{center}
\vspace{-4mm}
\caption{An example of the Spatio-Temporal And-Or Graph model, where the and-nodes represent compositions in either time or space, the or-nodes indicate structural alternatives, and the leaf-nodes (at the bottom) correspond to local part detectors. The links between leaf-nodes represent spatio-temporal contextual interactions.}
\label{fig:framework}
\end{figure}

To handle the first challenge, our STAOG model extends the traditional deformable graphical models by introducing switch variables in hierarchy, \emph{i.e.} or-nodes that explicitly specify structural reconfiguration. Both spatial and temporal interactions between action parts can be simultaneously learned. One action in the video can be treated as an ensemble of spatio-temporal compositions: a number of discrete temporal anchor frames, each of which is further decomposed into a layout of deformable parts. An example of the proposed STAOG model is illustrated in Fig.~\ref{fig:framework}. There are four layers in our model from bottom to top:

(1) The \emph{leaf-nodes} in the bottom layer represent a batch of local classifiers for detecting various action parts in every anchor frame, denoted by the solid circles in Fig.~\ref{fig:framework}. During detection, location displacement is allowed for each leaf-node to tackle the part deformation.

(2) The \emph{or-nodes} over the bottom are ``switch'' variables specifying the activation of their children leaf-nodes, denoted by the dashed circles. Each or-node is used to specify an appropriate selection from candidate action parts detected by the associating children leaf-nodes. In this way, by explicitly switching selections over leaf-nodes,
the or-nodes make our model reconfigurable during the inference of detection, which is the key to handle large action variabilities.

(3) The \emph{and-nodes} verify the holistic appearance of action within the anchor frame, (the rectangles in layer 2), and we thus consider it as the spatial and-node. It includes two aspects: (i) a global classifier with bag-of-features, and (ii) aggregated scores from its children or-nodes.

%(ii) spatial constraints (interactions) between action parts.

(4) The \emph{root-node} in the top can be viewed as an and-node in time, (the rectangle in top). Its definition is similar to the spatial and-node: (i) a classifier with global features in observed frames, (ii) aggregated scores over candidate temporal anchor frames, plus penalty for anchor frame displacements.

%plus temporal interactions between inter-frame action parts.

(5) The \emph{spatio-temporal contextual interactions}, \emph{e.g.} the curves (graph edges) among leaf-nodes in Fig.~\ref{fig:framework}, are defined based on informative contextual pairwise relations in either spatial or temporal domain. Note that the collaborative edges are imposed between leaf-nodes that are associated with different action parts. Their effectiveness will be particularly demonstrated in the experiments.

To overcome the second challenge, we present a novel weakly supervised learning algorithm for model training, inspired by the non-convex optimization techniques~\cite{cccp,NIPS12}. This algorithm trains the model in a dynamic manner: the model structure (\emph{e.g.} the configuration of leaf-nodes and or-nodes) is iteratively generated and reconfigured on the training data, with optimizing the multi-layer parameters. The other structure attributes (\emph{e.g.} the activation of leaf-nodes, and temporal deformation of anchor frames) are modeled with the latent variables and optimized simultaneously.

In the testing stage, we present an algorithm of cascaded search and verification for recognizing actions with the trained STAOG model. We first generate a set of hypotheses in both spatial and temporal compositions. (i) Spatial testing via the and-nodes. Within the input frame, all candidate action parts are found by leaf-nodes and several possible configurations (\emph{i.e.} spatial compositions of action parts) are produced with different specifications via the or-nodes. These configurations are also weighted via the and-node. (ii) Temporal testing via the root-node. The scores proposed by the and-nodes are aggregated via the root-node for a possible temporal composition. Several possible configurations are then produced as hypotheses represented by different latent variables. Finally, each hypothesis is globally verified with the spatial and temporal edges in the model.

This paper is organized as follows. Section 2 gives a brief review of related work. Then we present our STAOG model in Section 3, followed by a description of the inference procedure in Section 4. Section 5 presents a description for structural learning of our model. The experimental results and comparisons are exhibited in Section 6. Section 7 concludes this paper.

\section{Related work}
Traditional works for action recognition focused on developing informative features, such as spatio-temporal descriptors~\cite{STIP,DollarVSPETS05cuboids,zhu09},3D Gradient~\cite{3Dgradients}, 3D SIFT descriptors~\cite{SIFT} and motion features~\cite{TrajectoryModeling12,trajYan}, and the action classifier can be trained with labeled data. Most of these methods, however, are limited to periodic actions with clean background, such as running and jogging.

To address complex actions with cluttered background, several compositional or expressive models were proposed and achieve very impressive results~\cite{actioncontext,CrossDataset, actionbank,HMM,highlevel12,contextedges}. For example, Wang et al.~\cite{HMM} modeled the human action by a flexible constellation of parts conditioned on image observations and learned the parameters of an HCRF model in a max-margin framework, motivated by the recent progresses in object recognition and detection, \emph{e.g.} the deformable part model by Felzenszwalb et al.~\cite{iccp}. Yao et al.~\cite{ActionTemplate09} proposed to generate spatio-template action templates with the information projection method. Sadanand et al~\cite{actionbank} adopted high-level representations with a bank of individual action detectors. However, actions in video often involve much more information in both spatial and temporal domain, compared with image-based object recognition, and most of these studies do not explicitly localize parts of actions (actors) due to the computational burden. Moreover, structural configurations of these models are usually fixed, including a fixed number of part detectors as well as the predefined composition.

One unique characteristic of human action recognition problem lies in the temporal structure. A lot of works were proposed to build temporal structure models~\cite{HSMM05, ModelTemp,LatentTem_CVPR2012,SpatioTempGraph} based on discriminative and interesting motion segments of the video. Raptis et al.~\cite{DiscoverPart} extracted clusters of trajectories and proposed a graphical model to incorporate constraints for individual and group events. Albanese et al.~\cite{ProbabilisticPetriNet} represented temporal relations of activities using the probabilistic Petri Nets and integrated high-level reasoning approaches. Different from these approaches, we do not treat the whole temporal frames as units.Instead, we model temporal structure based on action parts with explicit relations and presents a solution to find both spatial and temporal configurations for dynamic activities.

Recently, the And-Or graph models~\cite{ZhuM06} have been discussed for several vision tasks such as object recognition~\cite{LinGrammar} and shape modeling~\cite{objectAndOrTree}. These works mainly focused on images instead of videos and do not take the temporal dynamic structure into account. The very recent work by Amer et al.~\cite{costsenstive} proposed to recognize activities with the spatio-temporal And-Or graph model, but they over-simplified the model training by manually fixing the model structure (\emph{i.e.} the layout of graph nodes).

It is worth mentioning that this paper learn the spatio-temporal graph without using any extra annotations or scripts. Research works which utilize rich  annotation for event parsing and interpretation are beyond the scope of this work. In contrast, Marszalek et al.~\cite{actioncontext} explored the action contexts of natural dynamic scenes with movie scripts. Gupta et al.~\cite{Storyline09} proposed to learn a visually grounded storyline model from annotated videos, and Pei et al.~\cite{ZhuEventAnd} studied the event grammar model for daily activities based on a predefined set of unary and binary relations. Extra annotations are required for these studies.

\section{Spatio-Temporal And-Or Graph}

The STAOG model is defined as $\mathcal{G = (V,E)}$, where $\mathcal{V}$ represents the four types of nodes and $\mathcal{E}$ the graph edges as Fig.~\ref{fig:framework}. The root node in top verifies the temporal composition, which aggregates scores over anchor frames. Each and-node represents a temporal anchor frame for verifying spatial composition. The or-nodes are derived from each and-node, which are ``switch'' variables for specifying the activation of their children leaf-nodes. The number of leaf-nodes for each or-node is dynamically learned with an upper limit number $m$. For simplicity, we use $t= 1,\dots,T$ to index all and-nodes in the whole STAOG model, $i = 1,\dots,Z$ for or-nodes and $j = 1,\dots,n$ for leaf-nodes. We also index the child or-node of and-node $A_t$ as $i\in ch(t)$, and index the child leaf-node of or-node $U_i$ as $j \in ch(i)$. The spatio-temporal graph edges(\emph{i.e.} interactions) are defined between the leaf-nodes associated with different or-nodes. In this section, we describe two factors in detail: the spatio-temporal compositions, and the contextual interactions in both spatial and temporal domains.
%-------------------------------------------------------------------------
\subsection{Spatio-Temporal Compositions}
\label{sec::Spatial}

We employ Laptev's 3-D corner detector~\cite{STIP} to detect interest points in video sequences, and each interest point is described by HoG (histogram of gradient) and HoF (histogram of optical flow)~\cite{realisticLap}. Furthermore, we generate a dictionary of spatio-temporal interest points' descriptors, clustered by the k-means method in training stage. Given a video sequence $X$, we first equally divide it into $T$ temporal segments. The center frame in each video segment is chosen as an initial anchor frame.  Each anchor frame is further decomposed into a number of action parts. In our method, we define the action parts based on the video patch representation, \emph{i.e.} 3-D volumes spanning $\rho$ consecutive frames. Thus, for each anchor frame $I_t$, we observe a sequence of frames centered at $I_t$, and the sequence denoted as $\Lambda_t$ is treated as the input for anchor frame processing.

%$I_t$, we further identify a number of action parts, each of which can be viewed as a 3-D volume spanning $\rho$ consecutive frames from $I_t$.  Thus, we denote $\Lambda_t$ as a sequence of frames centerred at $I_t$,
%
%And we use the Bag-of-Words (BoW) histogram $\Lambda_t$ to describe action parts with the generated dictionary.
%
% The feature domain of the anchor frame $I_t$ is denoted as $\Lambda_t$. %The spatial compositions of graph nodes in are presented as follows. %By using the equally sampled segments, we implicitly capture the fact that a primitive action occupies a roughly similar interval and is composed of similar number of frames. %

\begin{figure}[ptb]
\begin{center}
\epsfig{figure=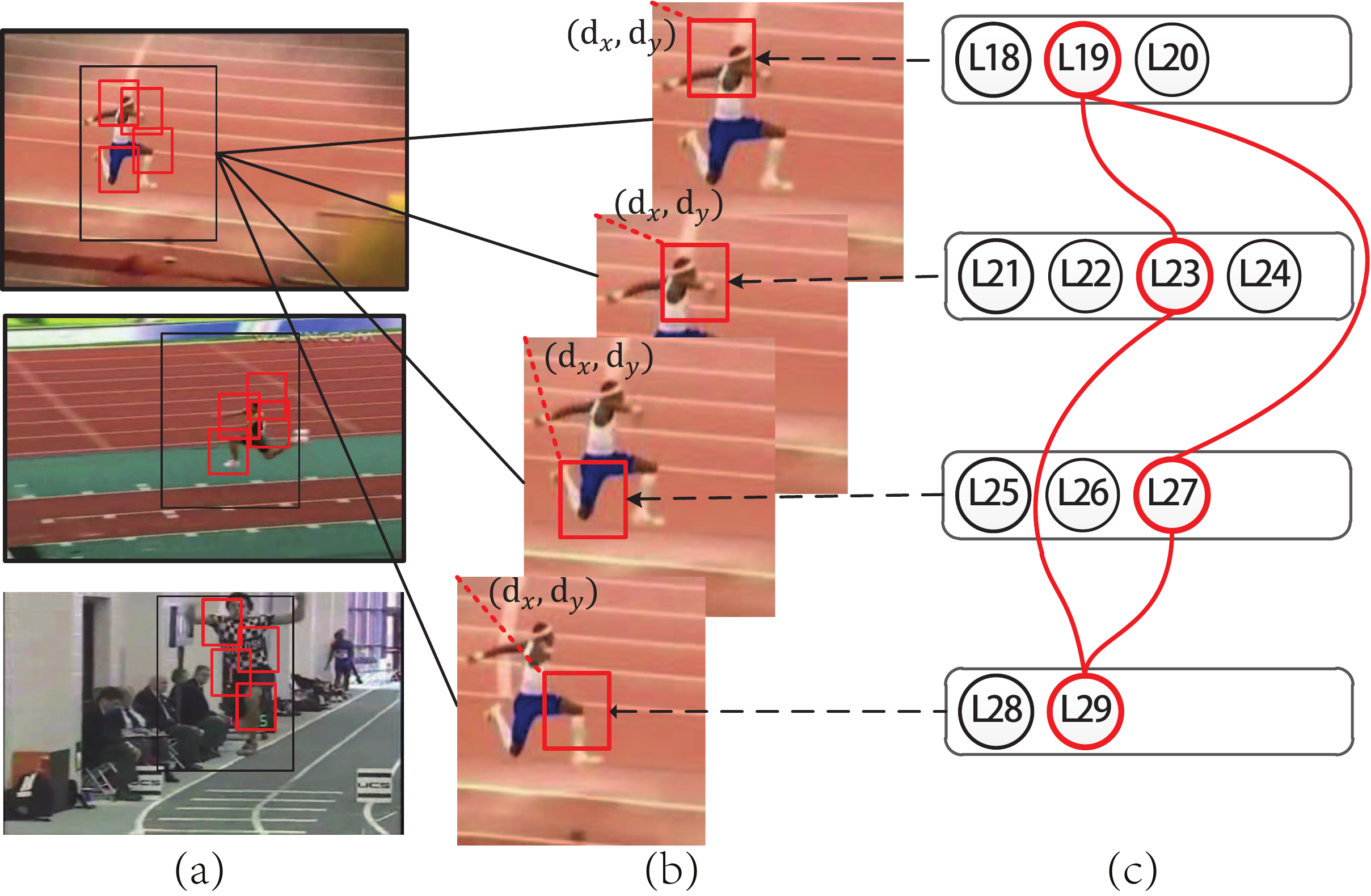,width=3.4in}
\end{center}
\caption{Illustration of spatial compositions. (a) The black boxes denote the initial positions of action parts. (b) The parts are exhibited which are associated with a set of leaf-nodes. Each $(d_x,d_y)$ indicates the location displacement determined by the model. (c) The activated leaf-nodes are highlighted by red and spatial contextual interactions are defined between the pairwise spatial adjacent leaf nodes.}
\label{fig:spa}
\end{figure}

\textbf{Leaf-node:} Each leaf-node $L_j$ represents a local classifier for detecting action parts within video patches and includes two terms: an appearance feature $\phi^l$ and the spatial displacement feature $\phi^s$. Within an anchor frame $I_t$, the features of action parts are described by the BoW histogram based on the generated dictionary.  Assume the action part detected by $L_j$ is localized at position $p_j =(p_{j}^x,p_{j}^y)$ , then $\phi^l(\Lambda_t,p_j)$ is denoted as the appearance feature. During detection, the locations of action parts are allowed to be perturbed to tackle the spatial deformation. We incorporate the spatial displacement $\phi^s(q_{j},p_{j}) = (d_x,d_y)$ for each action part, which can be computed by maximizing the response of $L_j$ during inference; $q_j$, representing the initialized position of $L_j$, is set according to the center-point of the frame. Thus the spatial displacement is defined as $\phi^s(q_{j},p_{j}) = (d_x,d_y)$, where $(d_x,d_y)$ is the displacement.The response of $L_j$ is defined as,

\begin{equation}\begin{aligned}
    R_{j}^l(\Lambda_t,p_j) = \omega_j^l\cdot\phi^l(\Lambda_t,p_j) - \omega_j^s\cdot\phi^s(q_{j},p_{j}),\label{eq:leafnode}
\end{aligned}\end{equation}
%\vspace{-2ex}
where $\omega_j^l$ is the parameter for the appearance feature and $\omega_j^s$ corresponds to the spatial deformation parameter.

\textbf{Or-node:} Each or-node $U_i$ is proposed to specify an appropriate candidate from its children leaf-nodes. For each leaf-node $L_j$ of $U_i$'s children, the indicator variable $v_j \in \{0,1\}$ represents whether it is activated or not and each or-node only selects one leaf-node. Briefly, we utilize the indicator vector $\mathbf{v}_i$ for the or-node $U_i$ and each element of $\mathbf{v}_i$ is an indicator variable $v_j$ of the leaf-node $L_j$. Intuitively, the significant intra-class variance caused by views, background clutters or actors can be captured by different spatial configurations that are determined with the or-nodes. The response of the or-node $U_i$ is defined as,

\begin{equation}\begin{aligned}
    R_{i}^u(\Lambda_t, \mathbf{v}_i)=\sum_{j\in ch(i)}R_{j}^l(\Lambda_t,p_j)\cdot v_j,\label{eq:ornode}
\end{aligned}\end{equation}

\textbf{And-node:} Each and-node $A_t$ verifies the holistic appearance of action for the anchor frame $I_t$, and spatial composition of the or-nodes in its children. We define the configuration vector $V_t$ for all leaf-nodes within the anchor frame, which includes all indicator vectors $\mathbf{v}_i$ corresponding to its children or-nodes $U_{i}$. The response of the and-node $A_t$ is defined as,

\begin{equation}\begin{aligned}
   \! R^a_t(\Lambda_t,V_t) =&\omega^a\cdot\phi^a(\Lambda_t) + \sum_{i\in ch(t)}R_{i}^u(\Lambda_t,\mathbf{v}_{i}),\label{eq:andnode}
\end{aligned}\end{equation}
where $\phi^a(\Lambda_t)$ is the BOW histogram globally extracted from the 3-D volume $\Lambda_t$ centered at $I_t$. The second term aggregates the response scores from all or-nodes of $A_t$'s children.

%The activation of leaf-nodes of or-nodes' children within the anchor frame are determined by the spatial composition response and global verification of the and-node $A_t$.
%-------------------------------------------------------------------------
%\subsection{Temporal Compositions}

%The best matching scores from each and-node are accumulated to model the temporal composition for the action.

\textbf{Root-node:} The root-node is a global potential function that verifies the temporal compatibility of model, including three terms: the global BoW histogram of the video clip, aggregated scores of its children and-nodes, and temporal displacements of anchor frames. Fig.~\ref{fig:temp} illustrates the temporal composition by the root-node. We employ the root-node for searching for the best localizations of $T$ anchor frames. We introduce the latent variable $\Delta_t$ to indicate the temporal displacement of each anchor frame $I_t$, which will be calculated during inference.  This implicitly carries the temporal ordering constraints which are crucial for discriminating human activities.

In particular, the temporal displacement penalty $\xi_t$ punishes the position of the and-node $A_t$ (\emph{i.e.} one anchor frame) shifting far away from the initial anchor point $\tau_t$ in time. Once $\Delta_t$ is optimized, the position of each anchor frame can be determined by $\tau_t + \Delta_t$ accordingly. We define $\xi_t$ by,
\begin{equation}\begin{aligned}
    \xi_t = -\omega_t^{\tau}\cdot\Delta_t,
\end{aligned}\end{equation}
where $\omega_t^{\tau}$ is the corresponding parameter. The response of the root-node can be then defined as,
\vspace{-3mm}
\begin{equation}\begin{aligned}
    \!\!R^r(X,\mathbf{V},\mathbf{\Delta})\!=\omega^r\cdot\phi^r(X) \!+ \sum^T_{t=1}R^a_t(\Lambda_t,V_t) + \xi_t,\label{eq:root}
\end{aligned}\end{equation}
where $\phi^r(X)$ is the BoW histogram feature extracted from the whole video sequence $X$.  $\mathbf{V} = (V_1,\cdots,V_T)$ and $\mathbf{\Delta} = (\Delta_1,\cdots,\Delta_T)$ are latent variables in the model for specifying the spatial and temporal configurations.

\begin{figure}[ptb]
\begin{center}
\epsfig{figure=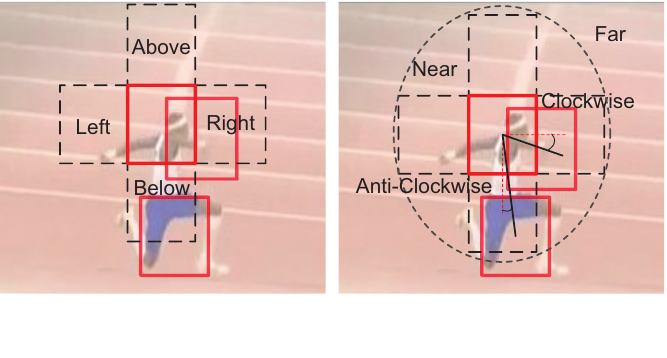,width = 3.2in}
\end{center}
\vspace{-4ex}
\caption{Illustration of the contextual relations for defining spatial edges in the STAOG model. We define the edges between spatial adjacent leaf-nodes with $8$ relations according to their spatial layout: above, below, left, right, near, far, clockwise and anti-clockwise.}
\label{fig:SpaInteraction}
\end{figure}

\subsection{Contextual Interactions}
\label{sec:context}

\textbf{Spatial interactions.} We impose spatial contextual interactions, \emph{i.e.} spatial edges, between pairwise spatially adjacent leaf-nodes in each anchor frame, as Fig.~\ref{fig:spa}(c) illustrates. Note that we only link the edges between a pair of leaf-nodes that are respectively associated with two different or-nodes.

For one edge connecting two leaf-nodes $(L_j,L_{j'})$, we define it with a set of informative relations, \emph{i.e.} a 8-bin binary feature $\varphi^s(L_j,L_{j'})$: \emph{above}, \emph{below}, \emph{left}, \emph{right}, \emph{near}, \emph{far}, \emph{clockwise} and \emph{anti-clockwise} between two adjacent leaf-nodes. The relations are visualized in Fig.~\ref{fig:SpaInteraction}. Suppose one edge connects two leaf-nodes $(L_j,L_{j'})$ which detect action parts at positions $p_j$ and $p_{j'}$ respectively. The centered red rectangle represents the location $p_j$, and the other red rectangles represent the adjacent parts. In the right chart of Fig.~\ref{fig:SpaInteraction}, the dashed line represents the initial layout of the two leaf-nodes, and the black solid line the adjusted actual layout during inference. Then we define the relations as,

\begin{itemize}
\setlength{\itemsep}{1pt}
\setlength{\parskip}{0pt}
 \setlength{\parsep}{0pt}
    \item \emph{near} or \emph{far}: If $p_{j'}$ is fallen into the outer dashed ellipse, it is near to $p_j$, \emph{i.e.} the bin of near is activated (\emph{i.e.} set as 1); otherwise it is far to $p_j$.\\
    \item \emph{above}, \emph{below}, \emph{left} or \emph{right}: The corresponding bin is set as 1 only if the center of $p_{j'}$ is inside the corresponding dashed rectangles.\\
    \item \emph{clockwise} or \emph{anti-clockwise}: one of the two relations is activated (\emph{i.e.} set as 1) according to the angle between the dashed line and the black solid line.\\
\vspace{-2ex}
\end{itemize}
The relations intuitively encode the spatial contexts of two action parts detected via the two leaf-nodes with respect to two different or-nodes. The response of the pairwise potentials can be parameterized as,
\vspace{-2mm}
\begin{equation}\begin{aligned}
    \Gamma^s_{jj'} = \beta^s_{jj'}\cdot\varphi^s(L_j,L_{j'}),\label{eq:spaI}
\end{aligned}\end{equation}
where $\beta^s_{jj'}$ is the corresponding 8-bin parameter vector.

\begin{figure}[ptb]
\begin{center}
\epsfig{figure=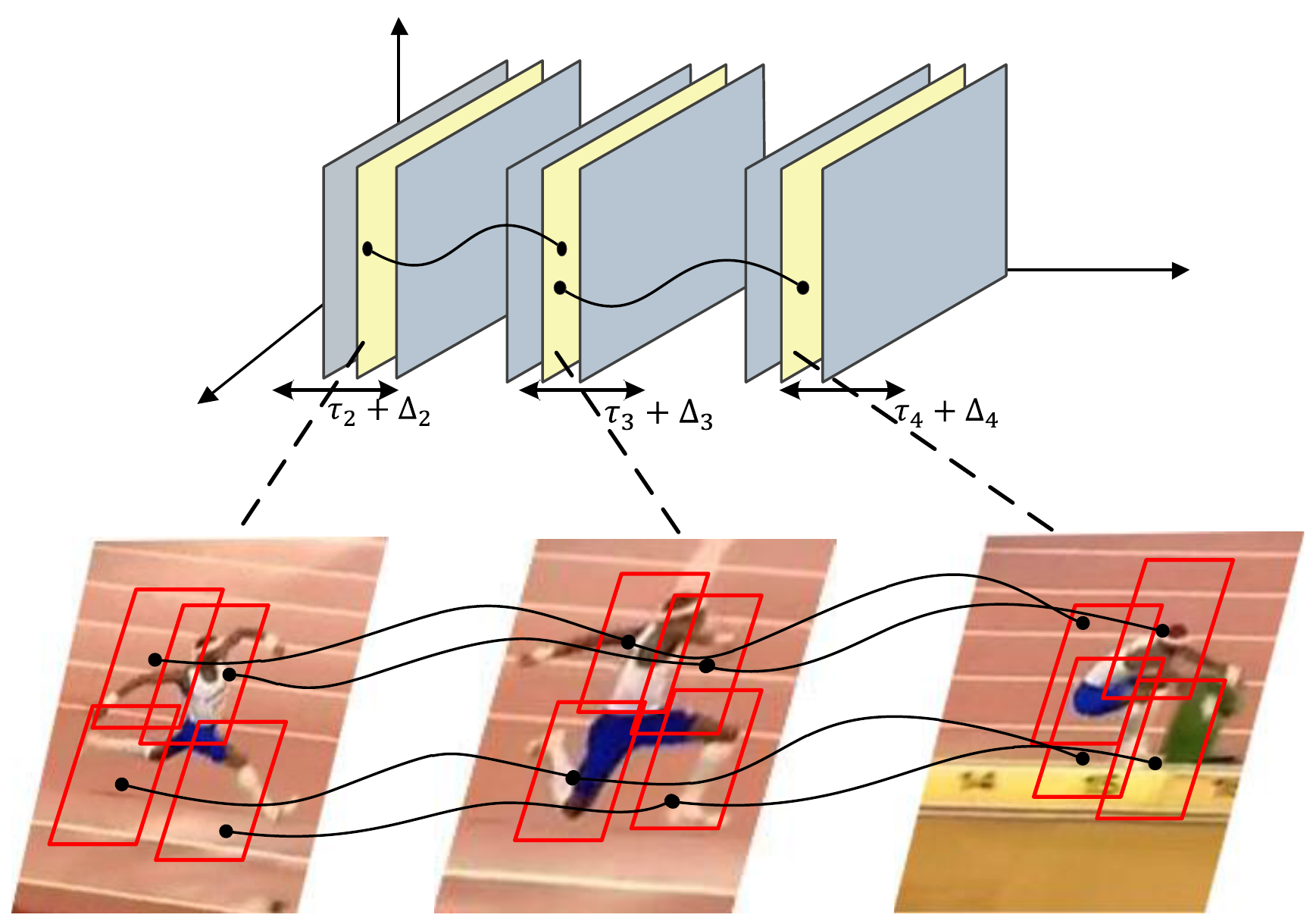,width=3.4in}
\end{center}
\vspace{-3mm}
\caption{Illustration of temporal compositions. The input video is decomposed into a number of discrete temporal anchor frames. The optimal position of each anchor frame is localized in $\Delta_t + \tau_t$, \emph{i.e.} the temporal displacement $\Delta_t$ plus the initial anchor point $\tau_t$. The temporal contextual interactions are defined in temporally adjacent action parts.}
\label{fig:temp}
\end{figure}

\textbf{Temporal interactions.} We also impose the edges in temporal domain in our model to represent the temporal interactions of action parts. The edges connect temporally adjacent leaf-nodes, illustrated in Fig.~\ref{fig:temp}. The edges are connected between any pair of leaf-nodes $(L_j,L_{j'})$ that belong to the same part within the two adjacent anchor frames respectively. A set of temporal relations are collected to concatenate a $4$-bin binary feature vector $\varphi^{\tau}(L_j,L_{j'})$.%These relations are divided into two kinds of predicates: temporal predicates and spatial predicates.

Specifically, we adopt four predicates: \emph{intersect}, \emph{after}, \emph{meets}, \emph{interrupt}, inspired by Allen's temporal predicates~\cite{AllenF94}~\cite{Spatiotemporalrelationship}. These predicates describe relations between two time intervals. The action part detected by one leaf-node $L_j$ for a specific anchor frame is described by the feature $\phi^l$ extracted from a 3-D volume with time span $\rho$.  Note that we ignore some predicates of ordering such as \emph{before} and \emph{equals}, as the order of temporally adjacent anchor frames is supposed to be fixed. Assume that leaf-node $L_j$ is associated with the anchor frame localized at $\tau_t + \Delta_t$, (initial position plus displacement). The starting and ending time $(f_{L_j}^{start}, f_{L_j}^{end})$ of $L_j$ can be calculated as,

\vspace{-3ex}
\begin{equation}\begin{aligned}
    f_{L_j}^{start} &= \tau_t + \Delta_t - \frac{\rho}{2},\\
    f_{L_j}^{end} &= \tau_t + \Delta_t + \frac{\rho}{2}.
\end{aligned}\end{equation}
Then we can define the four temporal predicates for the two temporal adjacent leaf-nodes $(L_j,L_{j'})$ as,
\begin{equation}\begin{aligned}
    \emph{intersect}(L_j, L_{j'}) &\Longleftrightarrow f_{L_{j'}}^{start} < f_{L_j}^{end},\\
    \emph{after}(L_j, L_{j'}) &\Longleftrightarrow f_{L_j}^{end} < f_{L_{j'}}^{start} < f_{L_j}^{end} + \rho,\\
    \emph{meets}(L_j, L_{j'}) &\Longleftrightarrow f_{L_{j'}}^{start} = f_{L_j}^{end},\\
    \emph{interrupt}(L_j, L_{j'}) &\Longleftrightarrow f_{L_j}^{end}+ \rho< f_{L_{j'}}^{start}.
\end{aligned}\end{equation}
Thus, we define the response of one temporal edge linking two leaf-nodes accordingly,

\begin{equation}\begin{aligned}
    \Gamma^{\tau}_{jj'} = \beta^{\tau}_{jj'}\cdot\varphi^{\tau}(L_j,L_{j'}),\label{eq:temI}
\end{aligned}\end{equation}
where $\beta^{\tau}_{jj'}$ is the corresponding 4-bin parameter. If the pairwise leaf-nodes $(L_j,L_{j'})$ satisfies the specific predicate, the corresponding bin is set to 1, otherwise 0.

% The two types of interactions are both constructed according to the co-occurrence of leaf-nodes. The response of the pairwise contextual interactions is parameterized as,
%
%\begin{equation} \begin{aligned}
%    R_{jj'}^e(L_j,L_{j'}) = -\kappa\cdot\log P(L_j,L_{j'}),\label{eq:interaction}
%\end{aligned} \end{equation}
%%\vspace{-3ex}
%where $P(L_j,L_{j'})$ denotes the probability of the co-occurrence of the leaf-node $(L_j,L_{j'})$ for all positive instances, and $\kappa$ is a tuning parameter set empirically.

Therefore, the overall response of the STAOG model is:
\begin{equation}\begin{aligned}
    R^g(X,\mathbf{V},\mathbf{\Delta})&= R^r(X,\mathbf{V},\mathbf{\Delta})+ \sum^n_j(\sum_{j'\in \gamma^s(j)}\Gamma^s_{jj'}\cdot v_j\cdot v_{j'}\\&+ \sum_{j'\in \gamma^{\tau}(j)}\Gamma^{\tau}_{jj'}\cdot v_j\cdot v_{j'}),\label{eq:graph}
\end{aligned}\end{equation}
%\vspace{-2ex}
where $(\mathbf{V},\mathbf{\Delta})$ are the hidden variables in the STAOG model. The second term defines the spatio-temporal contextual interactions between the leaf-nodes. $\gamma^s(j)$ is denoted as the set of leaf-node $L_j$'s neighbors which are spatially adjacent to the leaf-node $L_j$, and $\gamma^{\tau}(j)$ is introduced for the leaf-nodes which are temporally adjacent to the leaf-node $L_j$. Intuitively, spatial interactions between action parts guarantee the spatial coherence, as well as temporal interactions embedding the temporal contextual relations. Briefly, we refer $\mathcal{L} = (\mathbf{V},\mathbf{\Delta})$ as the latent variables in the following. The Eq.\ref{eq:graph} can be briefly written as:
\begin{equation}\begin{aligned}
    \!R^g(X,\mathcal{L}) = \psi\cdot\Phi(X,\mathcal{L}),\label{eq:global}
\end{aligned}\end{equation}
%\vspace{-4ex}
where $\psi$ includes the complete parameters of the STAOG model, and $\Phi(X,\mathcal{L})$ denotes the overall feature vector.
%\begin{equation} \begin{aligned}
%    \!\omega = &(\omega_1^l,\cdots,\omega_n^l,\omega_1^s,\cdots,\omega_n^s,\\
%    &\omega_1^a,\cdots,\omega_n^a,\omega_1^\tau,\cdots,\omega_n^\tau,\omega^r),
%\end{aligned} \end{equation}
%-------------------------------------------------------------------------

\section{Inference}

%The hypotheses based on temporal structure and spatial deformation for a given video instance are significant uncertain.
The inference task is to detect $T$ optimal temporal anchor frames for one video instance as well as the spatial composition of action parts within each anchor frame. In our approach, we perform a cascaded search that integrates three steps: spatial testing, temporal testing and global verification to maximize the global potential $R^g(X,\mathcal{L})$ defined in Eq.\ref{eq:graph}.

\textbf{Step 1. Spatial Testing via the and-nodes.}

The subgraph of the STAOG model, rooted at one and-node, can be viewed as the spatial composition classifier for localizing action parts in one frame. We first use all existing leaf-nodes to search for candidate actions parts. Assume the leaf-node $L_j$  associated with the frame $I_t$ detects the action part at the position $p_j^*$ by maximizing the response in Eq.\ref{eq:leafnode}. Each or-node is allowed to activate only one leaf-node, then a possible configuration consisting of action parts is decided by the indicator variables of the or-node, (\emph{i.e.} $\mathbf{v_i}$ for or-node $U_i$, indicating which leaf-node is activated). In this way, a set of possible configuration hypotheses $\{V_t\}$ are generated for further testing, which ensemble the hypotheses proposed by the or-nodes for the frame $I_t$. In practice, we limit the maximum number of hypotheses by setting a threshold on $R^a_t(\Lambda_t,V_t)$ in Eq.\ref{eq:andnode}.

\begin{small}
\begin{algorithm}[htb]
\caption{Inference Algorithm}
\label{alg:Inference}
\begin{algorithmic}\footnotesize
\INPUT ~~\\
    A learned STAOG model $G$, the action parts detected by all leaf-nodes by maximizing the response in Eq.\ref{eq:leafnode}.
\INITIAL ~~\\
    The set of possible hypotheses $l_t = \{\}$ for all $t\in{1,\dots T}$ anchor frames.
\INTERATION~~\\
\FORALL {$t = 1 \cdots T$ }
    \STATE  For each and-node $A_t$, a set of temporal displacement steps $\Sigma$ is predefined for sliding the possible anchor frames.
        \FORALL {$\Delta_t \in \Sigma$ }
        \STATE
        \begin{itemize}
        \setlength{\itemsep}{1pt}
        \setlength{\parskip}{0pt}
        \setlength{\parsep}{1pt}
           \item[(a)] initialize the set of pair terms $Q = \{Q_1,\dots,Q_K\}$ for all or-nodes of $A_t$'s children.
           \item[(b)] generate a set of pair terms $Q_i$ for each or-node $U_i$.
            \FORALL {$U_i,i \in ch(t)$}
                \FORALL{$L_j,j\in ch(i)$}
                \STATE %pool the leaf-node $L_j$ into $l^s_t$.\\
                    $Q_i = Q_i \cup (i,j)$.
                \ENDFOR
            \ENDFOR~~\\
            \item[(c)] obtain possible hypotheses $V_t = (\mathbf{v}_1,\dots,\mathbf{v}_K)$ by assembling the indicator variables of $K$ or-nodes according to the set $Q$.
            \item[(d)] The set of possible hypotheses for each specific displacement $\Delta_t$ is constructed as $l_t = l_t \cup (\{V_t\},\Delta_t)$.
        \end{itemize}

        \ENDFOR
\ENDFOR~~\\
Assemble these hypothesis \{$l_t\}$ for all $T$ anchor frames orderly to generate the set of hypotheses sequence $l$. Each possible configuration $(\mathbf{V},\mathbf{\Delta})$ belongs to the set $l$. The global response of STAOG model can be calculated by Eq.\ref{eq:final}.
\OUTPUT ~~\\
    \STATE
    The latent variables $\mathbf{V},\mathbf{\Delta}$ and the final score $S_\psi(X)$.
\end{algorithmic}
\end{algorithm}
\end{small}

\textbf{Step 2. Temporal Testing via the root-node.}

We apply the spatial testing with the and-nodes to localize a number of candidate anchor frames over several frames. The scores over candidate anchor frames (proposed by the and-nodes) are aggregated via the root-node for a possible temporal composition. For efficiency, we utilize a fixed number of discrete steps $\Sigma$ for searching each anchor frame. Several possible hypotheses are then produced with different anchor frame determinations by sliding the discrete steps $\Sigma$. In addition, we re-weight the hypotheses at the root-node in Eq.\ref{eq:root} by considering the temporal displacements of anchor frames as well as the the global features over the video clip. Intuitively, the hypotheses are represented by the specified latent variables $(\mathbf{V},\mathbf{\Delta})$.

\textbf{Step 3. Global Verification.}

Given all the hypotheses from the root-node, we apply the global potential function defined in Eq.\ref{eq:graph} to validate the optimal detection. The objective of the global verification is to cope with the noisy local detections on leaf-nodes. It combines the score of the root-node with the responses of spatial and temporal contextual interactions (edges).

The optimal response $S_\psi(X)$ of the model as well as the latent variables $(\mathbf{V},\mathbf{\Delta})$ can be calculated as,
\begin{equation} \begin{aligned}
S_\psi(X) = \mathop{max}_{\mathbf{V},\mathbf{\Delta}}(\psi\cdot\Phi(X,\mathbf{V},\mathbf{\Delta})).\label{eq:final}
\end{aligned} \end{equation}

Algorithm \ref{alg:Inference} summarizes the overall algorithm of the inference.

%-------------------------------------------------------------------------
\section{Structural Learning}

We formulate the structural learning of STAOG model as a joint optimization task for model structure and parameters. We solve this model by a novel latent learning method extended from the CCCP framework~\cite{cccp}. This algorithm iterates to train the model in a dynamic manner: the leaf-nodes can be automatically created or removed to reconfigure the model structure. The model structure is determined by latent variables $\mathcal{L} = (\mathbf{V},\mathbf{\Delta})$ that are inferred in each step.
\begin{figure*}[ptb]
\begin{center}
\epsfig{figure=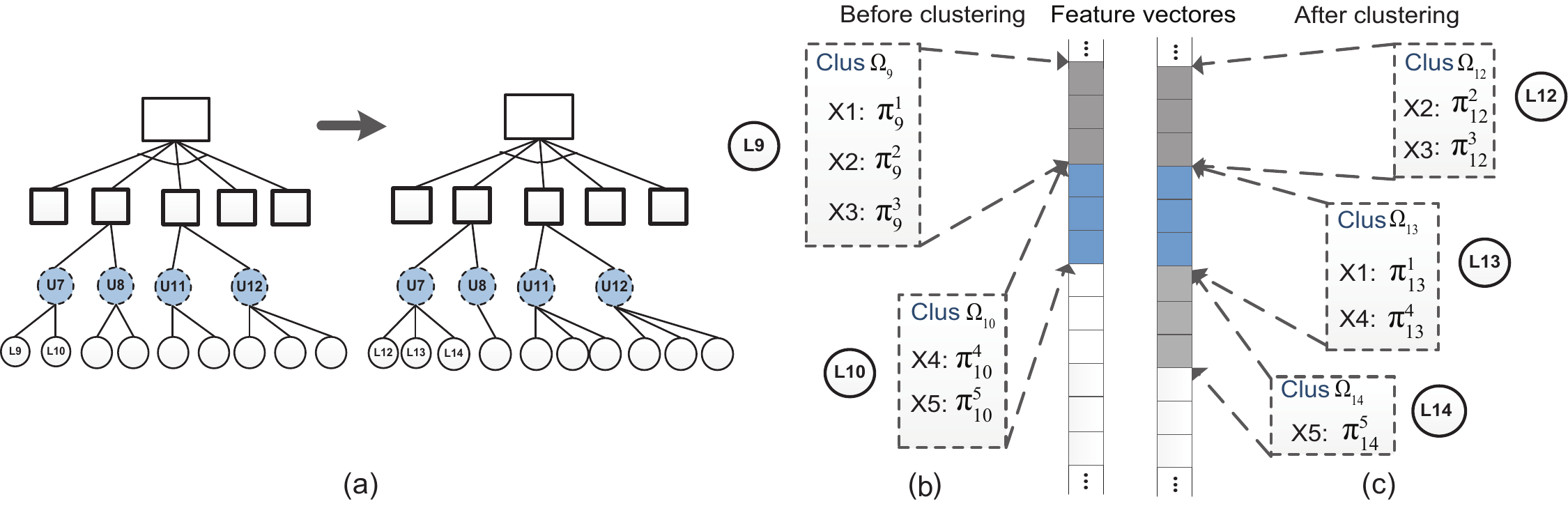,width = 1\textwidth}
\end{center}
\vspace{-4mm}
\caption{Illustration of discriminative structural learning. We reconfigure the model structure by re-arranging the feature vector, as the example illustrated. Parts of the STAOG model reconfigured in two iterations are shown in (a), where the left one represents the original model and the other one the new model. During this step, the new leaf-node associated with $U_7$ and $U_{11}$ are created and a leaf-node associated with $U_8$ is removed. Assume that we use 5 samples, $X_1,\dots,X_5$, for the structure learning. (b) shows the feature vectors detected by the same leaf-node are first grouped into one cluster, \emph{i.e.} one cluster for one leaf-node. (c) illustrates the feature rearrangement after clustering. For example, the feature vector  of sample $X_1$ is grouped from cluster $\Omega_9$ into cluster $\Omega_{13}$, we move the feature bins $\pi_9^1$ into the bins corresponding to the leaf-node $L_9$. Cluster $\Omega_{14}$ is a newly generated cluster, we thus create a new leaf-node accordingly.}
\label{fig:learning}
\end{figure*}

Let $D = ((X_1,y_1),(X_2,y_2),\dots,(X_N,y_N))$ be a set of labeled training samples, with $y_{k} \in \{1,-1\}$. The feature vector for each sample $(X,y)$ is defined as,
\begin{equation} \begin{aligned}
    \Phi(X,y,\mathcal{L}) = \begin{cases}
                            \Phi(X,\mathcal{L})& \text{if $y=+1$},\\
                            0& \text{if $y=-1$}.
                            \end{cases}
\end{aligned} \end{equation}
%\vspace{-2ex}
The temporal anchor frames and spatial configurations for each frame can be optimized by maximizing $R^g(X,\mathcal{L})$ in the inference procedure. We refer to $\mathcal{L} = (\mathbf{V},\mathbf{\Delta})$ as the latent variables, then we redefine Eq.\ref{eq:final} as,
\begin{equation} \begin{aligned}
    S_\psi(X) = \mathop{max}_{y,\mathcal{L}}(\psi\cdot\Phi(X,y,\mathcal{L})),
\end{aligned} \end{equation}
%\vspace{-3ex}
The optimization of this function can be solved by the latent structural SVM framework,
\begin{equation} \begin{aligned}
   \mathop{min}_\psi\frac{1}{2}\|\psi\|^2&+C\sum_{k=1}^N\lbrack \mathop{max}_{y,\mathcal{L}}(\psi\cdot\Phi(X_k,y,\mathcal{L}) + h(y_k,y))\\&- \mathop{max}_{\mathcal{L}}(\psi\cdot\Phi(X_k,y_k,\mathcal{L}))\rbrack,\label{eq:cost}
\end{aligned} \end{equation}
%\vspace{-3ex}
where $C$ is a penalty parameter set as 0.003 empirically and $h(y_k,y)$ is the cost function, where $h(y_k,y) = 0$ if $y_k = y$, otherwise 1. The optimization problem described above is not convex in general. Following the CCCP framework, we convert the function in Eq.\ref{eq:cost} into a convex and concave form as,
\begin{equation} \begin{aligned}\label{eq:cccp}
 &\mathop{min}_\psi\lbrack\frac{1}{2}\|\psi\|^2 + C\sum_{k=1}^N \mathop{max}_{y,\mathcal{L}}(\psi\cdot\Phi(X_k,y,\mathcal{L}) + h(y_k,y))\rbrack\\
   &- \lbrack C\sum_{k=1}^N \mathop{max}_{\mathcal{L}}(\psi\cdot\Phi(X_k,y_k,\mathcal{L}))\rbrack\\
   &= \mathop{min}_\psi\lbrack f(\psi) - g(\psi)\rbrack,
\end{aligned} \end{equation}
%\vspace{-3ex}
where $f(\psi)$ represents the first two terms, and $g(\psi)$ the last term. This leads to an iterative learning algorithm that alternates estimating model parameters and the hidden variables $\mathcal{L}$. However, we still need to dynamically determine the graph configuration, \emph{i.e.} the production of leaf-nodes associated with or-nodes. An additional step for dynamically reconfiguring structure is added between the two original steps. The procedure is presented as follows.

\begin{small}
\begin{algorithm}[htb]
\caption{Learning algorithm for STAOG model}
\label{alg:Framwork}
\begin{algorithmic}\footnotesize
\INPUT ~~\\
  Training samples,$D = ((X_1,y_1),(X_2,y_2),\dots,(X_N,y_N))$.
\OUTPUT ~~\\
  The trained STAOG model .
\INITIAL ~~\\
\STATE
\begin{itemize}
\setlength{\itemsep}{1pt}
\setlength{\parskip}{0pt}
 \setlength{\parsep}{1pt}
     \item[1.] Initialize the positions of action parts and anchor frames for all samples.
     \item[2.] Initialize the latent variables $\mathcal{L}$ and parameters $\psi$.
\end{itemize}
\MYWHILE
    \STATE
    \begin{itemize}
\setlength{\itemsep}{1pt}
 \setlength{\parskip}{0pt}
 \setlength{\parsep}{1pt}
 \item[1.] Estimate the latent variable $\mathcal{L}^*_k = (\mathbf{V}^*_k,\mathbf{\Delta}^*_k)$ for each positive sample $(X_k,y_k)$ during inference.
 \item[2.]
         \begin{itemize}
        \setlength{\itemsep}{1pt}
        \setlength{\parskip}{0pt}
        \setlength{\parsep}{1pt}
           \item[(a)] Localize the anchor frames and action parts using the current latent variables $(\mathbf{V}^*,\mathbf{\Delta}^*)$.
           \item[(b)] For each or-node $U_i$, we obtain a set of feature vectors of its children leaf-nodes for positive samples, and regroup the feature vectors by Spectral clustering.
           \item[(c)] Reconfigure leaf-nodes according to clustering results to generate a new structure. Calculate the energy Eq.~\ref{eq:opt} as $E(\psi_t^d)$.
         \end{itemize}

  \item[3.] \begin{itemize}
          \setlength{\itemsep}{1pt}
        \setlength{\parskip}{0pt}
        \setlength{\parsep}{1pt}
  \IF{$E(\psi_t^d)<E(\psi_t)$}
        \STATE
        Accept the new model structure and estimate the parameters $\psi_{t+1}=\mathop{argmin}_\psi(f(\psi+\psi\cdot q_t^d))$.
  \ELSE
        \STATE
        keep the previous model structure. Estimate the parameters $\psi_{t+1}=\mathop{argmin}_\psi(f(\psi+\psi\cdot q_t))$.
  \ENDIF
  \end{itemize}
    \end{itemize}
\MYENDWHILE
 The optimization function defined in Eq.~\ref{eq:cost} converges.
\end{algorithmic}
\end{algorithm}
\end{small}

(I) The model parameter $\psi_t$ obtained in the previous iteration is fixed. We find a hyperplane $q_t$ to upper bound the concave part $g(\psi)$ in Eq.~\ref{eq:cccp}. Specifically, $q_t$ is the derivative of $g(\psi)$. Thus, we have $-g(\psi)\leq-g(\psi_t) + (\psi-\psi_t)\cdot q_t,\forall\psi$. The optimal latent variable $\mathcal{L}_k^*$ is calculated by $\mathcal{L}_k^* = \mathop{argmax}_{\mathcal{L}}(\psi_t\cdot\Phi(X_k,y_k,\mathcal{L}))$ for each positive example. Then the hyperplane is constructed as $q_t= - C\sum_{k=1}^N\Phi(X_k,y_k,\mathcal{L}_k^*)$.

(II) In the second step, the STAOG model is adjusted by structural reconfiguration with applying the current model on training examples. The reconfiguration is performed for each part, \emph{i.e.} or-node, independently, with the fixed latent variable $\mathcal{L}^*_k = (V^*_k,\Delta^*_k)$. Note that each action part detected by one leaf-node is mapped to several feature bins at specific positions in the vector $\Phi(X_k,y_k,\mathcal{L}_k^*)$.

For each or-node $U_i$, we apply its children leaf-nodes for detecting parts in all positive samples. Assume that leaf-node $L_j$ detects an action part on the $k$-th sample using the feature vector $\pi_j^k$, which is a sub-vector of the complete feature vector of $k$-th sample, $\Phi(X_k,y_k,\mathcal{L}_k^*)$. And we obtain a set of feature vector $\{ \pi_j^k \}$ for all samples. The vectors detected by the same leaf-node are first grouped into one cluster, \emph{i.e.} one cluster for one leaf-node. We denote the cluster for the $j$-th leaf-node as $ \Omega_j$. Then we perform the spectral clustering algorithm with the Euclidean distance on vectors of all leaf-nodes of or-node $U_i$'s children for all positive samples, and the similar vectors are grouped together. We re-arrange the feature vectors for all samples based on the newly generated partition. For example, if the feature vector $\pi_j^k$ is grouped from $ \Omega_j $ into another cluster $ \Omega_j'$, we adjust the position of $\pi_j^k$ in $\Phi(X_k,y_k,\mathcal{L}_k^*)$, \emph{i.e.} by moving the feature bins into the position representing $j'$-th leaf-node. If $\Omega_j' $ is a newly generated cluster, we thus create a new leaf-node accordingly. By analogy, we remove one leaf-node if few samples are grouped into the corresponding cluster. In this way, the structure of $U_i$ is reconfigured with the feature vector re-arrangement. In practice, we constrain the extent of structural reconfiguration, \emph{i.e.} only few leaf-nodes can be created or removed in one iteration. We present a toy example in Fig.\ref{fig:learning} for illustration. In Fig.\ref{fig:learning}(a), a leaf-node associated with the or-node $U_7$ is created to better handle the intra-class variance; A leaf-node is removed if there is another similar one, (\emph{e.g.} the leaf-node associated with the or-node $U_8$). The sub-vector of $\pi_9^1$ of sample $X_1$ is grouped from cluster $\Omega_{9}$ to cluster $\Omega_{13}$, then the feature bins are moved from $\pi_9^1$ to $\pi_{13}^1$ as Fig.\ref{fig:learning}(b) shows.

After this structure reconfiguration, we obtain the new feature vector for each sample, $\Phi^d(X_k,y_k,\mathcal{L}_k^*)$, and the hyperplane is re-calculated as $q^d_t= - C\sum_{k=1}^N\Phi^d(X_k,y_k,\mathcal{L}_k^*)$, accordingly.

(III) The newly generated model structure can be represented by the feature vector $\Phi^d(X_k,y_k,\mathcal{L}_k^*)$. The model parameters can be learned by solving $\psi_{t}^d=\mathop{argmin}_\psi(f(\psi+\psi\cdot q_t^d))$. The optimization task in Eq.\ref{eq:cost} becomes,

\begin{equation} \begin{aligned}
   & \mathop{min}_\psi\frac{1}{2}\|\psi\|^2 + C\sum_{k=1}^N\lbrack \mathop{max}_{y,\mathcal{L}}(\psi\cdot\Phi(X_k,y,\mathcal{L}) + h(y_k,y))\\
   &- \psi\cdot\Phi^d(X_k,y_k,\mathcal{L}_k^*))\rbrack.\label{eq:opt}
\end{aligned} \end{equation}
%\vspace{-2ex}
This is a standard structural SVM problem, which can be solved in the cutting plane method and Sequential Minimal Optimization. The energy Eq.\ref{eq:opt} can be calculated by $E(\psi_t^d) = f(\psi_t^d)-g(\psi_t^d)$. We accept the new model structure until $E(\psi_t^d)<E(\psi_t)$ and $\psi_{t+1} = \psi_{t}^d$. Otherwise, we keep the model structure as in the previous iteration and the parameter vector is calculated by $\psi_{t+1}=\mathop{argmin}_\psi(f(\psi+\psi\cdot q_t))$. Thus, we ensure that the optimization function will decrease in each iteration. We repeat the 3-step iteration until convergence. Algorithm \ref{alg:Framwork} summarizes the overall algorithm of learning a STAOG model. In the case of multi-class classification, we use a one-against-rest approach and select the class with the highest score.

\section{Experiments}

We test our STAOG model on two different action recognition databases: UCF YouTube~\cite{youtube} and Olympics Sports~\cite{ModelTemp}.  The video resolution is normalized to $320 \times 240$. The YouTube dataset contains $11$ action categories, which is challenging due to large variation in camera motion, object appearance, pose/view and large intra-class variability. We follow the standard setup using leave-one-out cross validation for a pre-defined set of 25 folds.Average accuracy over all classes is reported as performance measure. The Olympic Sports dataset consists of 16 different sports classes that contain complex motions going beyond simple punctual or repetitive actions. The challenges of Olympic sports arise from background clutters, viewpoints and complex sequence of primitive actions. Each action is performed only by a single actor and represents a temporal sequence of primitive actions (\emph{e.g.} triple-jumping, pole-vault and diving). We use the same train-test split setup and the average precision (AP) for each of the action classes as in ~\cite{LatentTem_CVPR2012}.

\subsection{Implementation}
\label{sec:imple}

\begin{small}
\begin{table*}
  \centering
  \begin{tabular}{| c | c | c | c | c | c | c | c | }
    \hline\hline
     & Liu et al~\cite{youtube}.
     & Zhang et al~\cite{SpatioTemporalPhrases}.
     & Ikizler-Cinbis et al~\cite{Ikizler-CinbisS10}.
     & Dense trajectories~\cite{denseTraj}.
     & Ours-1
     & Ours-2
     & Ours(full)\\
     \hline
    \hline
    b\_shoot & 53.0\% & \textbf{98.0}\% & 48.5\% & 43.0\% & 58.4\% & 62.0\% & 77.9\%\\
    bike & 73.0\% & 74.0\% & 75.2\% & \textbf{91.7\%} & 82.1\% & 87.3\% & 88.6\%\\
    dive & 81.0\% & 80.0\% & 95.0\% & \textbf{99.0\%} & 98.4\% & 98.6\% & 98.8\%\\
    golf & 86.0\% & 68.0\% & 95.0\% & {97.0\%} & 95.7\% & 95.3\% & \textbf{97.4\%}\\
    h\_ride & 72.0\% & 65.0\% & 73.0\%& 85.0\% & 81.3\% & 86.0\% & \textbf{88.0\%}\\
    s\_juggle & 54.0\% & 67.0\% & 53.0\% & 76.0\% & 66.0\% & 81.6\% & \textbf{82.2}\%\\
    swing & 57.0\% & 71.0\% & 66.0\% & \textbf{88.0\%} & 85.2\% & 84.1\% & 85.4\%\\
    t\_swing & 80.0\% & 68.0\% & 77.0\% & 71.0\% & 69.6\% & 80.0\% & \textbf{80.7\%}\\
    t\_jump & 79.0\% & 80.0\% & 93.0\% & 94.0\% & 90.2\% & {94.7\%} & \textbf{95.8\%}\\
    v\_spike & 73.3\% & 77.0\% & 85.0\% & 95.0\% & 89.7\% & 90.6\% & \textbf{96.4\%}\\
    walk & 75.0\% & 54.0\% & 66.7\% & 87.0\% & 85.2\% & 86.2\% & \textbf{87.4}\%\\
    \hline
    \hline
    Accuracy & 71.2\% & 72.9\% & 75.2 \% & 84.2\% & 82.0\%& \textbf{86.0\%} & \textbf{88.9\%}\\
    \hline
  \end{tabular}
  \caption{Accuracy per action class and average accuracy for all classes on the YouTube dataset~\cite{youtube}.}\label{tab:youtube}
\end{table*}
\end{small}

\begin{figure*}[ptb]
\begin{center}
\epsfig{figure=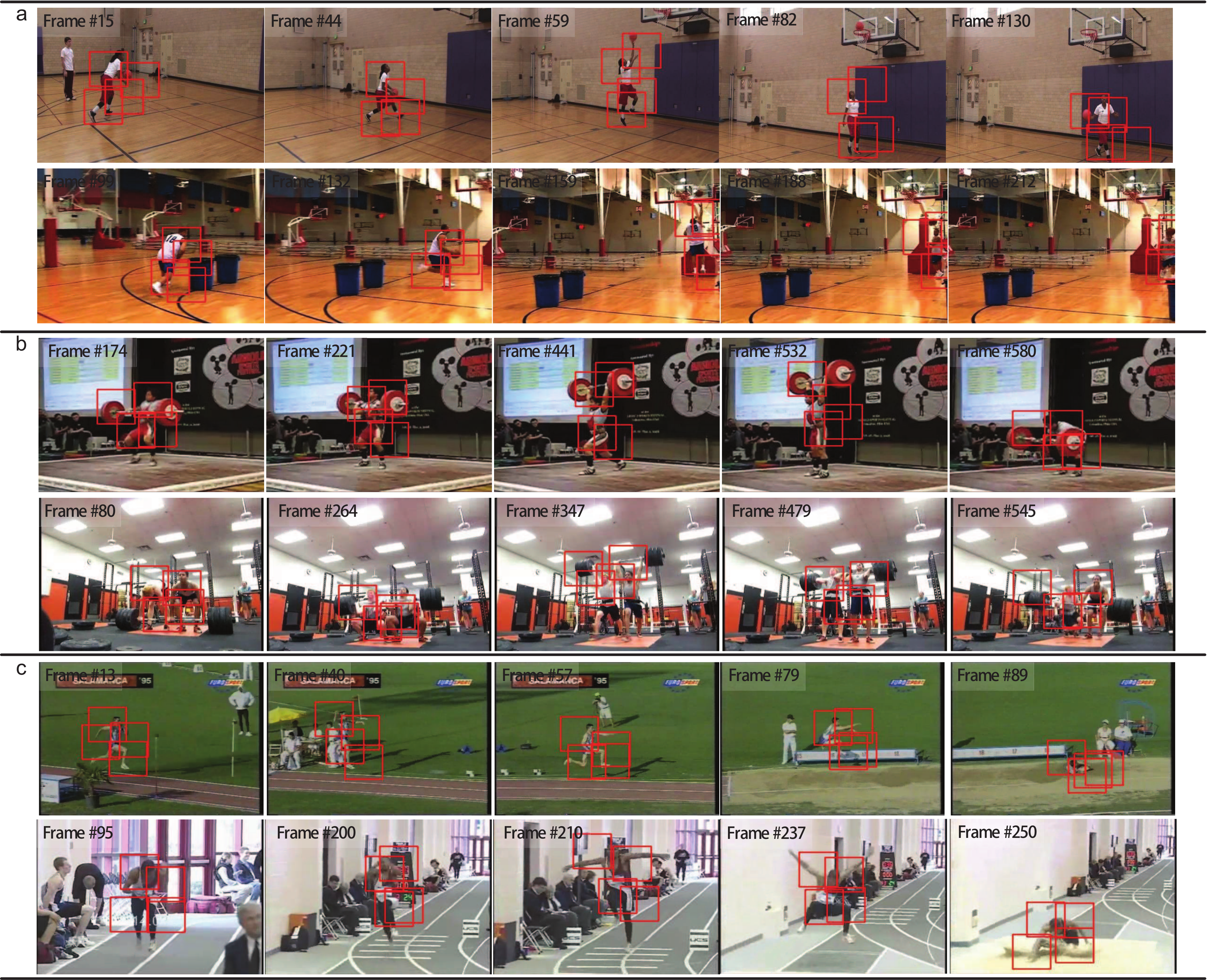, width =6.5in}
\end{center}
\vspace{-4mm}
\caption{Example inference results on three different action models \emph{e.g.} basketball-layup (a), clean-and-jerk (b) and triple-jump(c) learned on the Olympic Sports dataset and each of action category includes two instances. The red boxes in each frame represent the discovered discriminative action parts. Our model successfully localizes the accurate anchor frames across the instances in the long action videos. In addition, it is noticed that the large intra-class variabilities can be captured by our model.}
\label{fig:part}
\end{figure*}

We fix the number of and-nodes (anchor frames) in the STAOG model as $T = 3$ for UCF YouTube dataset, and $T = 5$ for Olympic Sports dataset empirically. We need more anchor frames for Olympic dataset because the actions are more complex and last over more frames. The parameter $T$ can be roughly estimated by the action temporal complexity in general. The number of spatial layout for each anchor frame is fixed as $2 \times 2$, and thus $K = 4$ or-nodes for each anchor frame. There are at most $m = 4$ leaf-nodes associated to each or-node. We extract the interest points described by the HOG and HOF features by utilizing the code published in~\cite{realisticLap} beforehand.  The size of one action part (within a 3-D volume) is empirically set to $60 \times 60$ pixels spanning $\rho=15$ frames. The dimension of the generated dictionary is set as $300$ for describing action parts in each anchor frame and the global features of the anchor frames. We set the discrete temporal steps searching for anchor frames: $\Sigma = [\pm{2}, \pm{4}, \pm{6}, \pm{8}, \pm{10}]$. The convergence of our learning algorithm usually takes $9\sim 10$ iterations.

\begin{small}
\begin{table}
  \centering
  \begin{tabular}{| c | c | c | c | }
    \hline\hline
     & Niebles et al~\cite{ModelTemp}.
     & Tang et al~\cite{LatentTem_CVPR2012}.
     & Ours(full)\\
     \hline
    \hline
    h-jump & 27.0\% & 18.4\% & \textbf{35.6\%}\\
    l-jump & 71.7\%& 81.8\% & \textbf{86.4\%} \\
    t-jump & 10.1\%& 16.1\% & \textbf{36.2\%}\\
    p-vault & \textbf{90.8}\% & 84.9\% & 84.3\% \\
    g-vault & \textbf{86.1}\% & 85.7\% & 83.1\% \\
    s-put & 37.3\% & 43.3\% & \textbf{56.8\%} \\
    snatch & 54.2\% & {88.6\%} & \textbf{89.0\%}\\
    c-jerk & 70.6\%& 78.2\% & \textbf{83.3\%}\\
    j-throw & \textbf{85.0}\% & 79.5\% & 78.1\%\\
    h-throw & 71.2\%& 70.5\% & \textbf{75.4\%}\\
    d-throw & 47.3\% & 48.9\% & \textbf{53.3\%}\\
    d-platform & \textbf{95.4}\% & 93.7\% & 92.8\%\\
    d-board & \textbf{84.3}\% & {79.3\%} &76.5\%\\
    basketball & 82.1\% & 85.5\% & \textbf{86.7\%}\\
    bowling & 53.0\% & \textbf{64.3\%} & 62.0\%\\
    t-serve & 33.4\% & 49.6\% &\textbf{62.3\%}\\
    \hline
    \hline
    mAP & 62.5\% & 66.8\% & \textbf{71.4\%}\\
    \hline
  \end{tabular}
  \vspace{-4mm}
  \caption{Average Precision(AP) values on the Olympic Sports dataset~\cite{ModelTemp}.}\label{tab:olym}
\end{table}
\end{small}

The experiments are carried out on a PC with Core I5 3.0GHZ CPU and 4GB memory. The average CPU-time used to process a video from Olympic Sports dataset is $200$ seconds and $150$ for a video in the UCF YouTube dataset. In particular, it takes 7 seconds for processing one frame in Olympic Sports dataset, and 4 seconds for each frame in the UCF YouTube dataset on average. The efficiency of our method is slightly diverse with the densities of the feature points in the video sequence.

\subsection{Results and Comparisons}
%\vspace{-1mm}
Compared with the recently proposed methods on the YouTube dataset, our model outperforms the state-of-the-art: we achieve the accuracy of 88.9\% in YouTube dataset, the reported results of the competing algorithms are: 71.2\% in~\cite{youtube}, 72.9\% in ~\cite{SpatioTemporalPhrases},75.2\% in~\cite{Ikizler-CinbisS10} and 84.2\% in~\cite{denseTraj}. The accuracy scores for all categories are reported in Table~\ref{tab:youtube}. Our method outperforms on $7$ out of the $11$ categories which have relatively large intra-class variance or background disturbance. In the Olympic Sports dataset, we obtain better AP scores for $10$ out of the $16$ categories, and overall AP score $71.4\%$, better than the previous methods~\cite{LatentTem_CVPR2012,ModelTemp}. The competing method proposed by Tang et al.~\cite{LatentTem_CVPR2012} utilizes the variable-duration HMM for learning the temporal structure in the video. Our results show that our compositional model with the explicit spatial and temporal relations can achieve better performance. The detailed results are reported in Table~\ref{tab:olym}.

Figure.~\ref{fig:part} illustrates the inference results on three action categories from the Olympic Sports dataset, each of which includes two instances. Our model also localizes the action parts in the anchor frames, as they are discriminative in appearance and motion. These results demonstrate well the capability of our model, since the scenarios of actions contain the very realistic challenges in video action recognition. The spatial compositions defined over the and-nodes enable us to handle pose/view variations and background disturbances. The temporal compositions are effective to localize anchor frames in videos against various motion frequency, temporal locations and video length.

For further evaluation, we conduct three empirical analysis in different model settings as follows.

(I) We simplify the temporal compositions by discarding the displacement $\Delta_t$ for each anchor frame, \emph{i.e.} fixing the temporal structure. We report AP scores of this model setting in the fourth column of Table~\ref{tab:youtube}, named as "Ours-1". The average accuracy is $82\%$,$6.9\%$ less than our complete model.

%In addition, our model obtains better performance in all categories by utilizing the temporal displacements.

(II) We also evaluate the benefit of spatio-temporal contextual interactions. Our model can be simplified into a tree structure by removing the interactions. The accuracies are shown in the sixth column of Table~\ref{tab:youtube}, named as "Ours-2". We can observe that the spatio-temporal contextual interactions make the accuracies increase $2.9\%$ on average. The increased performance of the interactions also speaks in favour for our model, as it shows that through better associations between the anchor frames and action parts, it is possible to achieve even better accuracies.

\begin{figure}[ptb]
\begin{center}
\epsfig{figure=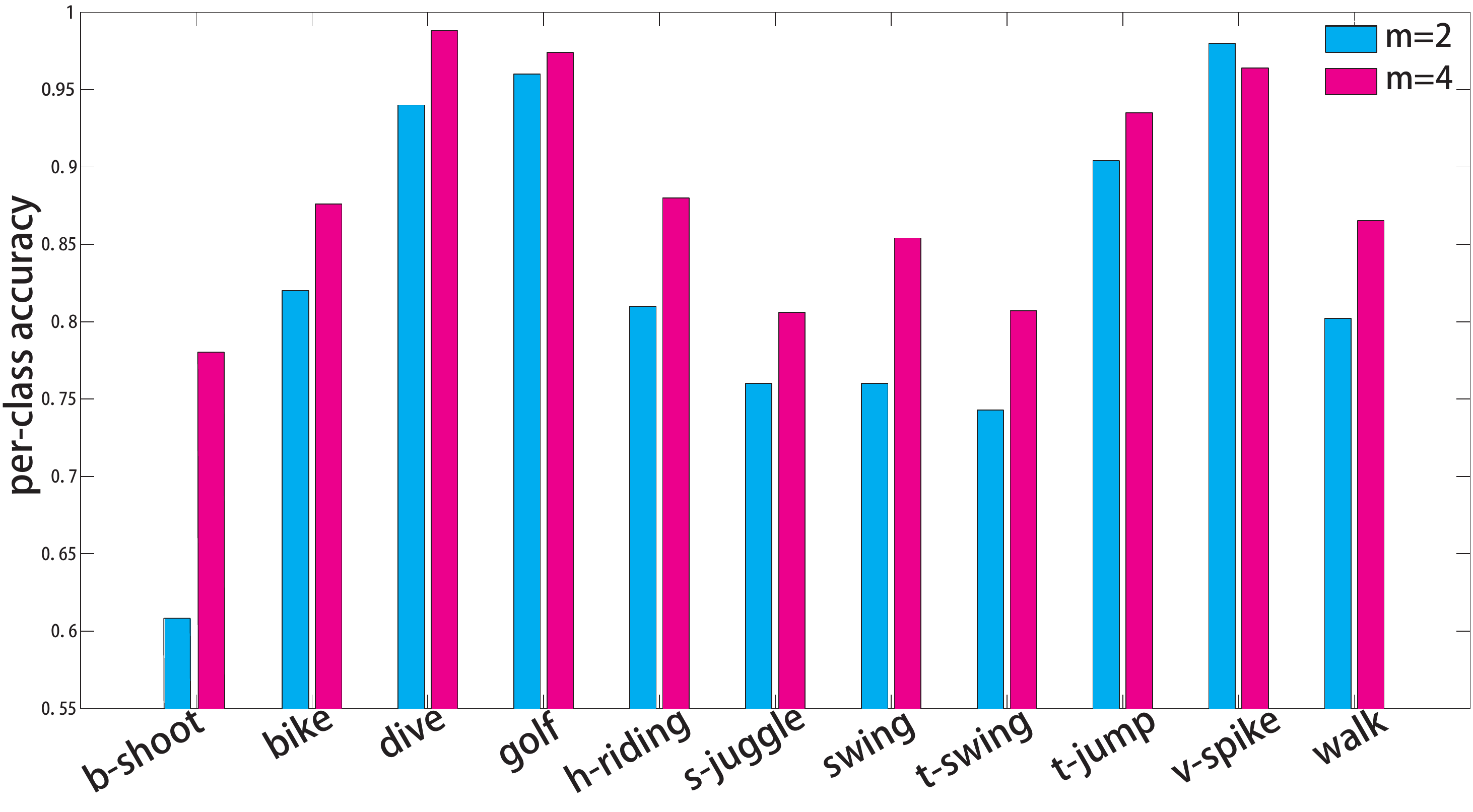, width = 3.4 in}
\end{center}
\vspace{-5ex}
\caption{Empirical analysis for different settings of spatial compositions, where we set different maximum numbers $m$ of leaf-nodes under the or-nodes. Each pillar represents the accuracy for one action category. The color indicates the results with different settings, $m = 2$ or $m = 4$.}
\label{fig:empirical_leafnodes}
\end{figure}

(III) One may be interested in how the performances are improved by introducing the or-nodes in spatial compositions, which is one of the key components in the STAOG model. In this experiment, we set different maximum numbers $m$ of leaf-nodes under the or-nodes, \emph{i.e.} how many leaf-nodes at most can be created for the model.  We compare the results on the YouTube dataset in Fig.~\ref{fig:empirical_leafnodes}, and observe that $m = 4$ achieves the better results in general than $m=2$, $82.4\%$. In practice, the number $m$ is not sensitive, as the exact number of leaf-nodes is decided by the clustering on data during the structural learning.

%(III) In addition, we present an empirical study to evaluate the benefit of spatial decomposition within anchor frames. For each or-node, we set different maximum numbers $m$ for its children leaf-nodes. That is, how many leaf-nodes at most can be created to account for action variance. It is actually a trade-off between model specialization and model complexity. In practice, we select $m = 4$ because it yields in good performance in all the experiments.

%\begin{figure}[ptb]
% \begin{center}
%\epsfig{figure=leafnodes.pdf, width = 3.4in}
%\end{center}
%\vspace{-4mm}
%\caption{Empirical analysis for spatial decomposition. In the STAOG model, we set %different maximum numbers $m$ of leaf-nodes for each or-node, and compare the %results. Each pillar represents the average-precision (AP) score for an action category. %The color indicates the results with different settings, $m = 2$ or $m = 4$.}
%\label{fig:empirical_leafnodes}
%\end{figure}

\section{Conclusion}
\vspace{-1mm}
This paper studies a novel hierarchical model for human action recognition, in the form of a configurable Spatio-Temporal And-Or Graph. This model is shown to handle well realistic challenges in action recognition. Moreover, we consider two aspects to improve our method. First, the model can be integrated with high level semantic information to represent multi-agent complex events. Second, we plan to speed up the algorithm for large-scale processing.

%includes incorporating the context and mutual interactions information into our model, which can help to understand more complex events such as human interactions, group activities and so on.
\vspace{-2mm}
\bibliographystyle{abbrv}
\bibliography{egbib}

%\bibliographystyle{ieee}
%\begin{thebibliography}{9}
%\balancecolumns % GM June 2007
% That's all folks!
\end{document}